# COMPARISON OF MACHINE LEARNING AND STATISTICAL APPROACHES FOR DIGITAL ELEVATION MODEL (DEM) CORRECTION: INTERIM RESULTS

Chukwuma Okolie[1,2,3*], Adedayo Adeleke[4], Julian Smit[5], Jon Mills[3], Iyke Maduako[6] and Caleb Ogbeta[7]

[1]Division of Geomatics, University of Cape Town, South Africa; oklchu002@myuct.ac.za
[2]Department of Surveying & Geoinformatics, University of Lagos, Nigeria; drcjokolie@gmail.com
[3]School of Engineering, Newcastle University, United Kingdom; jon.mills@newcastle.ac.uk
[4]Department of Geography, Geoinformatics and Meteorology, University of Pretoria, South Africa; adedayo.adeleke@up.ac.za
[5]Department of Civil Engineering and Geomatics, Cape Peninsula University of Technology, South Africa; smitj@cput.ac.za
[6]Department of Geoinformatics and Surveying, University of Nigeria, Nsukka, Nigeria; iykemadu84@gmail.com
[7]Geomatics Lab, School of Civil and Construction Engineering, Oregon State University, USA; ogbetac@oregonstate.edu

## 1. INTRODUCTION

Several methods have been proposed for correcting the elevation bias in digital elevation models (DEMs) for example, linear regression (e.g. Preety et al., 2022). Nowadays, supervised machine learning enables the modelling of complex relationships between variables, and has been deployed by researchers in a variety of fields. In the existing literature, several studies have adopted either machine learning or statistical approaches in the task of DEM correction. However, to our knowledge, none of these studies have compared the performance of both approaches, especially with regard to open-access global DEMs. Our previous work has already shown the potential of machine learning approaches, specifically gradient boosted decision trees (GBDTs) for DEM correction, e.g. (Okolie et al. 2023). In this study, we share some results from the comparison of three recent implementations of gradient boosted decision trees (XGBoost, LightGBM and CatBoost), versus multiple linear regression (MLR) for enhancing the vertical accuracy of 30 m Copernicus and AW3D global DEMs in Cape Town, South Africa.

## 2. METHODOLOGY

The training/input datasets are comprised of eleven predictor variables including elevation, slope, aspect, surface roughness, topographic position index, terrain ruggedness index, terrain surface texture, vector ruggedness measure, percentage bare ground, urban footprints and percentage forest cover. The target variable (elevation error) was derived with respect to highly accurate airborne LiDAR. Since multicollinearity is not a major concern for decision trees, all the input variables were fed into the gradient boosted decision trees (GBDTs) where training was done using Python scripting in the Google Collaboratory environment. Generally, the models (trained with default hyperparameters) performed considerably well and demonstrated excellent predictive capability. In the case of MLR, surface roughness and TRI were flagged during multi-collinearity (Person's correlation and Variance Inflation Factor) diagnostics and excluded from the input variables. Thus using MLR, the elevation error was expressed as a linear combination of nine input variables. The MLR was implemented within R, using the syntax for the lm() function. Both models (GBDTs and MLR) were evaluated at several implementation sites for prediction and correction of DEM error. The corrections were achieved by subtracting the predicted elevation errors from the original elevations (i.e., $DEM_{Corrected} = DEM_{Original} - \Delta h$).

## 3. RESULTS AND DISCUSSION

Numerous terrain offsets degraded the accuracy of the original DEMs. In several instances after correction, the terrain offsets in the original DEMs were de-escalated (e.g. Figures 1 and 2). Table 1 compares the percentage reduction in RMSE of AW3D and Copernicus DEMs after correction. In the urban/industrial and grassland/shrubland landscapes, there was a greater than 70% reduction in the RMSE of the original AW3D DEM, after correction. Similarly, the RMSEs reduced in other landscapes: agricultural (>45%), peninsula (>50%) and mountainous (>13%). The corrections improved the accuracy of Copernicus DEM, e.g., > 44% RMSE reduction in the urban area and >32% RMSE reduction in the grassland/shrubland landscape. The statistical-based (MLR) and machine learning (GBDT) correction achieved significant corrections of AW3D and Copernicus DEMs. While MLR outperformed the GBDTs in one scenario (i.e. Copernicus DEM in the grassland/shrubland landscape), the GBDTs outperformed MLR in most landscapes.

## 4. CONCLUSION

The comparison proves the robustness of the GBDT-based correction in virtually all the landscapes under consideration. Future studies could integrate other approaches in the comparison.

## ACKNOWLEDGEMENTS

The authors are grateful to the University of Cape Town and the Commonwealth Scholarship Commission for funding this research; and the Information and Knowledge Management Department, City of Cape Town for providing the LiDAR DEM.

*Corresponding author

| Landscape | % RMSE reduction (AW3D DEM) | | | | % RMSE reduction (Copernicus DEM) | | | |
|---|---|---|---|---|---|---|---|---|
| | MLR | XGBoost | LightGBM | CatBoost | MLR | XGBoost | LightGBM | CatBoost |
| Urban/ industrial | 72.1 | 72.2 | 72.8 | 72.5 | 44.3 | 46.8 | 46.7 | 47.0 |
| Agricultural land | 44.6 | 46.8 | 47.4 | 47.3 | 21.2 | 37.4 | 36.9 | 37.9 |
| Mountain | 13.4 | 17.4 | 16.9 | 17.6 | 6.2 | 10.0 | 9.8 | 10.3 |
| Peninsula | 51.7 | 52.2 | 52.5 | 51.9 | 8.4 | 14.1 | 14.6 | 16.1 |
| Grassland/ shrubland | 72.6 | 72.6 | 73.2 | 73.1 | 41.4 | 32.9 | 35.3 | 32.3 |

**Table 1.** Percentage reduction in RMSE of the original DEMs after correction

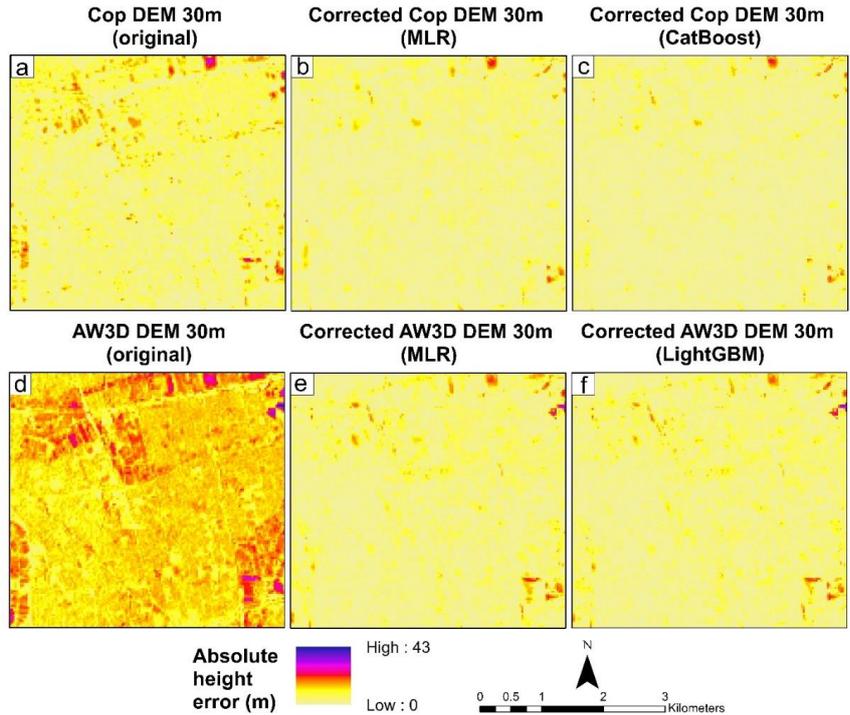

**Figure 1.** Absolute height error comparison of corrected DEMs in urban landscape

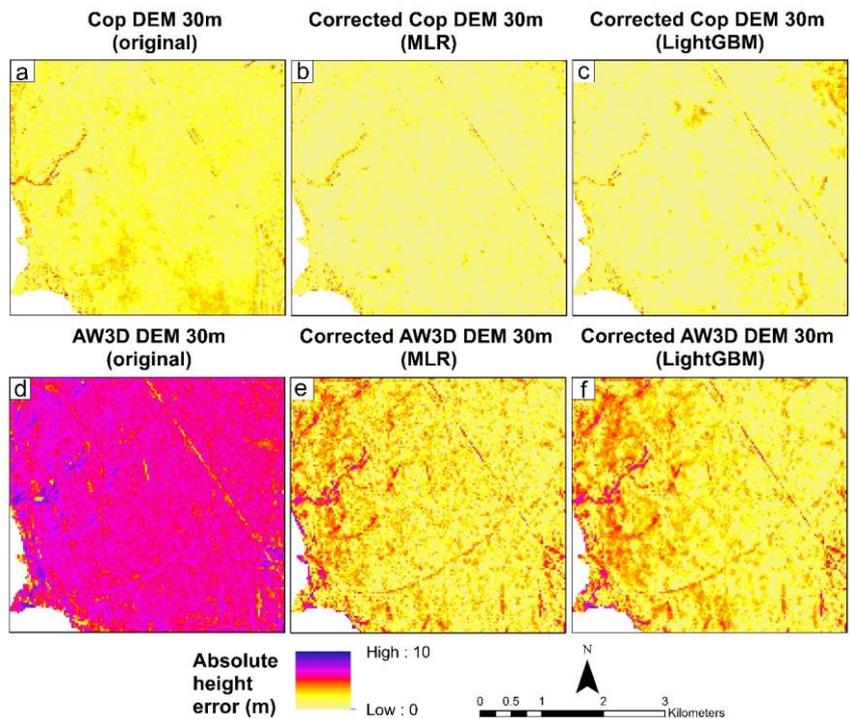

**Figure 2.** Absolute height error comparison of corrected DEMs in grassland/shrubland landscape